\documentclass{article}

\usepackage{PRIMEarxiv}

\usepackage[utf8]{inputenc} % allow utf-8 input
\usepackage[T1]{fontenc}    % use 8-bit T1 fonts
\usepackage{hyperref}       % hyperlinks
\usepackage{url}            % simple URL typesetting
\usepackage{amsfonts}       % blackboard math symbols
\usepackage{nicefrac}       % compact symbols for 1/2, etc.
\usepackage{microtype}      % microtypography
\usepackage{lipsum}
\usepackage{fancyhdr}       % header
\usepackage{graphicx}       % graphics
\graphicspath{{media/}}     % organize your images and other figures under media/ folder

\title{A Novel Evolutionary Algorithm for Hierarchical Neural Architecture Search}

\author{
  Aristeidis Christoforidis \\
  Dpt. of Applied Informatics \\
  University of Macedonia \\
  Thessaloniki, Greece\\
  \texttt{aid20003@uom.edu.gr} \\
   \And
  George Kyriakides \\
  Dpt. of Applied Informatics \\
  University of Macedonia \\
  Thessaloniki, Greece\\
  \texttt{ge.kyriakides@uom.edu.gr} \\
   \And
  Konstantinos Margaritis \\
  Dpt. of Applied Informatics \\
  University of Macedonia \\
  Thessaloniki, Greece\\
  \texttt{kmarg@uom.edu.gr} \\
}

\begin{document}
\maketitle

\begin{abstract}
In this work, we propose a novel evolutionary algorithm for neural architecture search, applicable to global search spaces. The algorithm's architectural representation organizes the topology in multiple hierarchical modules, while the design process exploits this representation, in order to explore the search space. We also employ a curation system, which promotes the utilization of well performing sub-structures to subsequent generations. We apply our method to Fashion-MNIST and NAS-Bench101, achieving accuracies of $93.2\%$ and $94.8\%$ respectively in a relatively small number of generations.
\end{abstract}

% keywords can be removed

\section{Introduction}
Neural Architecture Search is becoming an increasingly important sub-field of neural networks, able to produce state-of-the-art architectures without human intervention \cite{tanveer2021fine}. Among others, a number of evolutionary methods have been proposed \cite{Lyu2020_iym,9439793,Kriakides2021Evolving,Liu2020_rgc}, most utilize conservative mutation operators, where small changes of the underlying topologies are incurred. Although this allows the exploitation of well-performing topology regions, it can greatly hinder exploration of other regions. Furthermore, drastic changes to a topology can have destructive results in its performance. As such, a tool to sufficiently alter the structure of a network, while preserving well-performing features is needed. This is even more critical in global search spaces, where the search space can be unbounded, and thus novel architectural patterns can be discovered. Current methods perform better in cell-search spaces, were the macro-structure of the network is known beforehand, although this prohibits novel architectures from emerging.

Hierarchical search spaces attempt to find a middle ground between global and cell search spaces by conducting an augmented search at both levels. The micro search space contains cells with graphs comprised of neural layers, while the global search space has graphs where nodes are replaced by cells by the optimization algorithm.In \cite{Liang2018_fyx}, the authors update their DeepNEAT algorithm by restructuring the search space as hierarchical, on CoDeepNEAT, and apply an evolutionary algorithm to develop better populations in each set. At each evolution step they combine networks from both populations by replacing the nodes at the networks of the macro population with networks from the cell population to build the final networks. Their approach is pretty straightforward, with only two search levels.

In \cite{Liu2018_whp}, a more dynamic methodology is presented, proposing an algorithm that creates populations on arbitrary levels of up to a certain depth. Their modular approach allows the to tackle the problem of image classification efficiently, iteratively producing populations of performant networks trained on the CIFAR-10 dataset. After training with 200 GPUs for 1 hour they manage to produce a state of the art network with 3.6\% error, which is a 97\% decrease in search time.

In this work, we propose a novel evolutionary algorithm, which organises the generated networks in hierarchical modules. Contrary to similar approaches, such as \cite{Liu2018_whp, Liang2018_fyx}, we do not define a fixed number of hierarchical levels. Instead, we employ a recursive-like approach, allowing any hierarchical level to exist. We first describe the key concept and components of our method, which include the neural module, module lists and mutation operator. Following, we present experimental results on the Fashion-MNIST \cite{xiao2017fashion} and NASBench-101 datasets \cite{ying2019bench}. Finally, we discuss our findings, as well as possible future directions.

\section{Hierarchical Evolutionary Algorithm}

By combining an evolutionary approach with a novel hierarchical network representation, an unrestricted, iterative network architecture search algorithm that can produce deep neural networks of increasing complexity and performance can be built. The evolutionary process operates on a population of candidate network architectures that start simple and are expanded through mutation. The algorithm does not use the crossover operator for producing new offspring networks; even though it would be technically possible, it is very hard to implement a system compatible with the network representation that would guarantee an increasing trend in the fitness of the population. As such, the core components of the algorithm are the following:
\begin{itemize}
\item The neural module, which is an autonomous neural sub-structure, similar to a cell in cell-search spaces.
\item The notable, candidate, and banned module lists.
\item The mutation operator.
\end{itemize}

\subsection{Neural Modules}
\begin{figure}[t]
\centerline{\includegraphics{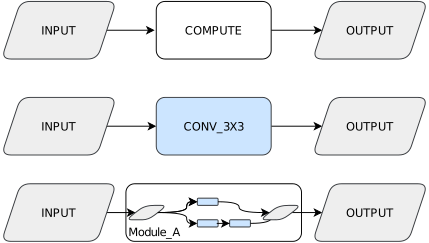}}
\caption{\textbf{Top:} a simple neural module with one computational node. \textbf{Middle:} the same module, where the computational node references a 3x3 Convolution Layer. \textbf{Bottom:} the same module, where the computational node references another module.}
\label{fig:neural_module}
\end{figure}

Neural modules are the basic building blocks of our networks, as well as members of the algorithm's population. Each neural module consists of an input and output node, as well as a number of computational nodes. Each computational node can either reference another neural module or a neural layer implementation (such as a convolutional or pooling layer). A simple example is depicted in Fig. \ref{fig:neural_module}. A neural module with a single computational node is depicted. At the top, the abstract computational graph is depicted. Each node is labelled as INPUT, COMPUTE, or OUTPUT, indicating that it is either an input, computational, or output node, respectively. In the middle, the implementation of the module is depicted, in the case where the computational node references a 3X3 convolution layer. Finally, the bottom depicts the implementation when the node references another complex module.

\begin{figure}[t]
\centerline{\includegraphics{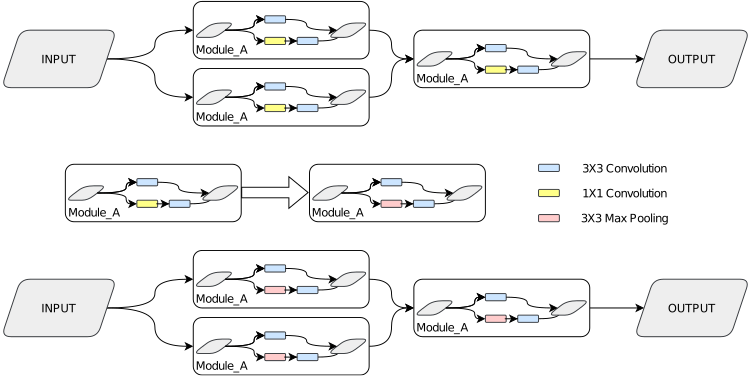}}
\caption{\textbf{Top:} a topology utilizing Module\_A, consisting of 3X3 and 1X1 convolutions. \textbf{Middle:} changing the 1X1 convolution with a 3X3 pooling operation. \textbf{Bottom:} the resulting network.}
\label{fig:neural_module_changes}
\end{figure}

By allowing a computational node to reference another module, it is possible to create complex networks. Furthermore, it is possible to drastically alter a network's topology in a controlled manner. For example, consider the top architecture in Fig. \ref{fig:neural_module_changes}. It consists of a repeating neural module, consisting of 3X3 and 1X1 convolution layers. By swapping the 1X1 convolution with a 3X3 pooling operation, we are able to change the architecture significantly in three distinct points. Furthermore, we do so in a controlled manner, i.e. we do not have to alter each occurrence of the layer individually. 

\subsection{Module Lists}
In order to promote well-performing modules while also avoid evaluating under-performing ones, we propose the utilization of a memory mechanism. This memory is implemented in the form of three lists:
\begin{itemize}
\item The notable modules list,
\item The candidate modules list, and the
\item and banned modules list
\end{itemize}

The notable modules list retains all generated modules that have historically performed best and their average fitness. The candidate modules list contains all generated modules currently under consideration for admission to the notable modules list. Finally, the banned modules list contains all generated modules rejected or removed from the notable modules list.

\subsection{Module Generation}
In order to generate a new neural module, the number of computational nodes \textit{N} in the neural module’s  graph is selected. This parameter can be set as a constant value, or be sampled from a distribution to add variance to the members of the population. During the first step of network generation, \textit{N} neural modules are sampled from the notable modules list, utilizing their average fitness as weights. Each selected neural module corresponds to a computational node in the new module. During the second step, a random computational directed graph is created for the new module, utilizing the \textit{N} computational nodes.

\subsection{Module Mutation}
The mutation operator, when applied to a neural module can either mutate one of its computational nodes, or one of its edges. Node mutation first chooses one computational node at random. If the node references a neural layer, it replaces its reference with a randomly generated neural module. If the node references a neural module, the operator is applied to the referenced module, in a recursive fashion. 

Edge mutation adds an edge between two nodes, either in the current module or in one of its children. First, the operator chooses at random if it will add the edge in the current module or one of its computational nodes. In the first case (adding the edge in the current module), the operator adds an edge between two random nodes, as long as the operation does not create a cycle in the graph. In the second case (adding the edge in a computational node), the operator chooses one computational node at random and repeats the process in the module referenced by the node.

\subsection{Module Evaluation and Lists Management}
Members of the population that are new or have been mutated at the current
iteration must be evaluated. This means that the corresponding networks have to
be constructed, trained and tested on the validation dataset. In order to evaluate a module, the corresponding network is constructed (according to the modules referenced by its computational nodes) and trained. As we would like to give more complex networks more time to train, we define the training epochs as:

\[training\_epochs=max(1,min(\frac{complexity}{max(log(generation+1),1)}, max\_epochs))\]

where max\_epochs is the maximum number of training epochs allowed and complexity is the number of computational nodes in the module.

Given the fitness value of a candidate in the population, all computational nodes share this fitness, independently of their complexity or depth. This is the same technique used by \cite{Liang2018_fyx}, although it is applied to a population of chromosomes with different structure. The idea is that good modules should consistently appear on well performing networks, therefore their fitness should not be penalized by their position or the depth in which they appear. However, an important problem may arise how will more complex modules be highlighted for their performance since all modules in a topology are assigned the same value? This issue is handled organically by the algorithm through the candidate and notable modules lists.

The notable modules list is initialized by creating and inserting into it a set of predefined neural modules, each representing a single neural layer (essentially graphs with one node), one for each layer available to the algorithm. Usually, convolutional and pooling layers with different properties make up the layer list. The first population members are created by combining these simple neural modules in different permutations, building slightly more complex neural modules. The new networks are trained and evaluated, and all the modules under the hierarchy are assigned fitness values. Modules that are already recorded in the notable modules list have their average fitness values updated.This initially concerns simple modules, like the modules representing neural layers that make up the computational graphs of the candidates. As for the new, higher level modules that have not yet been recorded in the notables list, they have to first be placed in the candidate modules list, where they will remain until it can be determined whether or not they are of quality.

\begin{figure}[t]
\centerline{\includegraphics{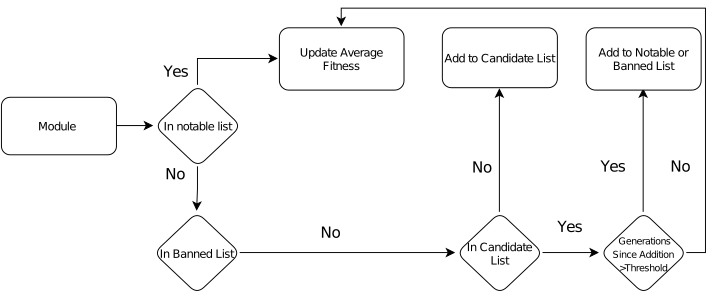}}
\caption{Decision process for neural modules admission to one of the three lists.}
\label{fig:list_mgmt}
\end{figure}

When a new module is first encountered in the population it is recorded in the candidate modules list. Each module in the candidate modules list must be evaluated a specific number of times before its average fitness is computed. Furthermore, this must be achieved in a specific time frame (limited number of generations), called time-to-leave (TTL). If the module fails to produce an average fitness better than the worst performing notable module within the allotted time, it is moved to the banned list. Modules in the banned list cannot be re-admitted to the candidate modules list. Fig. \ref{fig:list_mgmt} depicts the decision process.

\section{Experimental Results}
In order to evaluate our proposed methods, we conducted experiments on the NASBench-101 \cite{ying2019bench} and Fashion-MNIST \cite{xiao2017fashion} datasets, utilizing the NORD framework \cite{kyriakides2020nord}. All code is available on github, at \href{https://github.com/ArisChristoforidis/Dynamic-Hierarchical-NAS}{https://github.com/ArisChristoforidis/Dynamic-Hierarchical-NAS}. All experiments were conducted on Google's Collab platform, utilizing an NVIDIA Tesla K80 GPU.

\subsection{Fashion-MNIST}
The first dataset, Fashion-MNIST, contains 28 × 28 sized grayscale images of 10 different classes of articles of clothing and accessories. The classes are the following: t-shirt, trouser, pullover, dress, coat, sandal, shirt, sneaker, bag and ankle boot. There are 60000 images in the training set and 10000 images on the test set. In a attempt to speed up the search process, we define a set of 6 neural operations: 3 2D convolutional layers with an output channel count of 32 and kernel size of 1,2,3 and 3 max pooling layers with kernel size of 2,3,5. Convolutional layers are followed by a relu+batchnorm+dropout(with dropout probability equal to 10\%) layer group.

Our algorithm is configured to operate on a population of 20 modules (to adhere to temporal and size limitations), with a maximum size of 10 modules in the notable modules list and a base TTL of 4 for candidate modules. The minimum candidate topology observation count is 2 observations, with the fitness threshold activating at the end of every generation with the 40\% worst performing topologies in the population being replaced by new schemes. This last setting is fine tuned so that the well performing topologies can remain in the population and have a chance to expand their structure, improving their performance, while the ineffective networks are replaced completely, offering a drastic contribution to the candidates list. The node mutation chance is set to 15\% while the edge mutation chance is set to 55\%. At most, only one mutation may occur at a given topology per generation. This is implemented by sampling from a uniform distribution in the range [0, 1] and choosing the type of mutation based on the value of the sample s where s in [0, 0.15) corresponds to a node mutation, s in [0.15, 0.7) corresponds to an edge mutation and s in [0.7, 1] corresponds to no mutation. Each node mutation resulted in the generation of an abstract graph with two nodes.

\begin{figure}
\centerline{\includegraphics[width=0.8\textwidth]{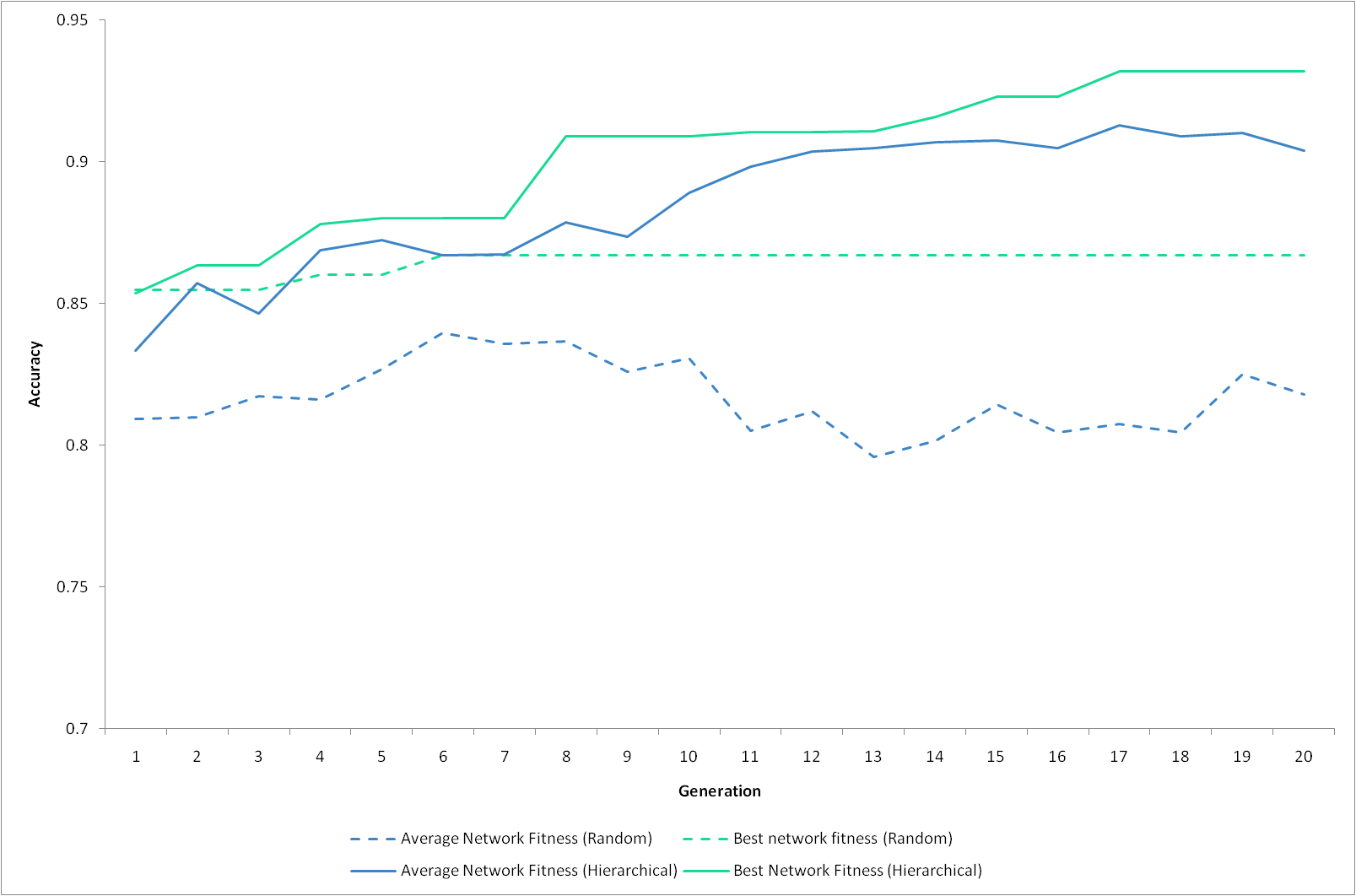}}
\caption{Our method vs. random search on the Fashion-MNIST dataset.}
\label{fig:fmnist}
\end{figure}

In order to test the effectiveness of our method we test it against random search (dashed lines). The random search is implemented by setting the same fitness value for all candidate topologies, regardless of the modules’ performance on the validation set. This gives all neural modules an equal chance of being promoted to notables, provided that they meet the occurrence threshold (and that the notables list is not full). We perform a limited architecture search for 20 generations and present the results in Fig. \ref{fig:fmnist}. 

Our approach manages to discover a topology with an accuracy score of 93.2\% in generation 17. As expected, the random search offers inferior results in both average network accuracy and best topology performance, achieving a top accuracy of 86.7\%. In fact, the average accuracy of the networks in the population on our approach surpasses the best performance of the random approach early on thanks to the sampling technique used in the notable modules. Our approach compounds knowledge from previous generations, which leads to the construction of better modules and networks possessing good predictive abilities. The accuracy score achieved using this approach is 3.71 percentage points from the state of the art score, achieved using a fine tuned DARTS based solution \cite{tanveer2021fine}. This is a gap that could theoretically be covered if the search runs for a higher number of generations or performing augmentation on the dataset. There is no sign of convergence in the curves, which indicates that a higher score is indeed possible.

\subsubsection{Generated Network Architecture}

\begin{figure}
\includegraphics[width=\textwidth]{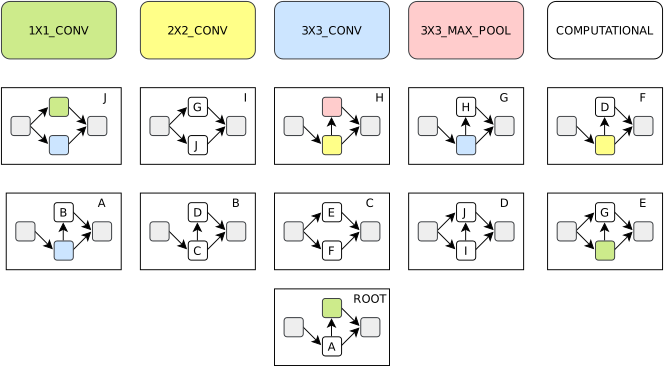}
\caption{Discovered neural modules on the Fashion-MNIST dataset.}
\label{fig:fmnist_arc}
\end{figure}

The best network topology uses 23 neural nodes and 93 connections between them. There are a total of 10 abstract modules used in the hierarchy, some of them repeating multiple times. The neural modules referenced in the final architecture are depicted in  Fig. \ref{fig:fmnist_arc}. It is interesting to note that the 23 nodes and 93 connections of the final architecture are achieved with only 9 distinct layer references, further highlighting how a small, controlled change in one of them can drastically change the final network's architecture.

\subsection{NASBench-101}
For our second experiment, we apply our dynamic hierarchical NAS algorithm to the CIFAR-10 dataset using the NAS-Bench-101 benchmark. While our approach is designed with global search spaces in mind, we aim to test how well it works on cell search spaces as well, where multiple restrictions apply. In this case, the search space comprises of cells with up to 7 nodes (neural operations) and 9 edges (connections), using only 3 available operations: 1×1 and 3×3 convolutions (grouped with an additional batch normalization and a RELU layers) and a 3×3 max pooling operation. There are 423624 possible cells in the search space. The discovered cell is placed in designated positions in a larger, fixed convolutional topology to form a full network.

\begin{figure}
\centerline{\includegraphics[width=0.7\textwidth]{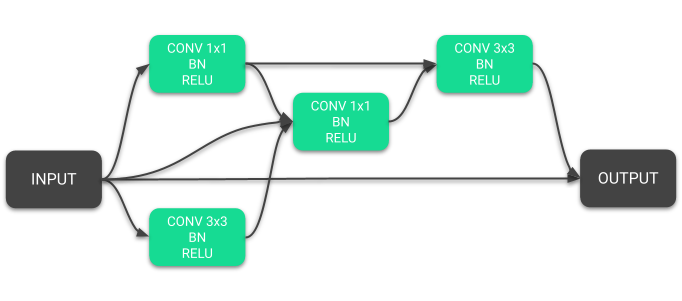}}
\caption{Discovered neural modules on the Fashion-MNIST dataset.}
\label{fig:nasbench_arc}
\end{figure}

Since there are significantly fewer operations in this experiment, a smaller population size of 10 is set. Candidate modules reach their occurrence threshold much faster, so new schemes are introduced quickly to the notable modules list. One thing that must be taken into account is the restrictions on the number of nodes and edges. If a candidate cell topology exceeds the imposed limits, it can’t be evaluated. Nevertheless, the search algorithm manages to find a top performing cell topology in just 6 generations with an accuracy of 94.8\%, depicted in Fig. \ref{fig:nasbench_arc}.

 \section{Conclusions}
In this paper, we propose a novel hierarchical evolutionary algorithm for Neural Architecture Search in global search spaces. The method utilizes blocks of re-usable neural modules, as well as three lists that aid in the management of the modules. By allowing module nodes to reference other modules, as well as through a mutation operator, the method is able to produce architectures of arbitrary complexity. Furthermore, the algorithm is able to explore the search space efficiently, as small, controlled changes in a module can have a profound effect on the generated architectures. We apply our methodology to two popular benchmark datasets. First, we apply it to the image recognition dataset, Fashion-MNIST, where the algorithm is able to generate a network with 93.2\% accuracy. Following, we apply the algorithm to the NAS benchmarking dataset, NASBench-101, where the algorithm is able to produce a cell with 94.8\% accuracy.

While the approach still suffers from some common problems present in most evolutionary algorithms, mainly the large amounts of time required to train and evaluate candidate networks such negative effects are mitigated in two ways. First, the population is able to converge relatively quickly, due to the controlled manner that major changes occur in the generated architectures. Second, the nature of the algorithm allows the utilization of predictive modelling concerning the performance of each candidate architecture. In future works, we aim to incorporate such predictive models in our search strategy, in order to further exploit the benefits of our method.
 
%Bibliography
\bibliographystyle{unsrt}  
\bibliography{references}

\end{document}